\documentclass[11pt,a4paper]{article}
\usepackage[latin9]{inputenc}
\usepackage{verbatim}
\usepackage{float}
\usepackage{mathtools}
\usepackage{url}
\usepackage{amsmath}
\usepackage{graphicx}
\usepackage{esint}

\makeatletter

\pdfpageheight\paperheight
\pdfpagewidth\paperwidth

\providecommand{\tabularnewline}{\\}
\floatstyle{ruled}
\newfloat{algorithm}{tbp}{loa}
\providecommand{\algorithmname}{Algorithm}
\floatname{algorithm}{\protect\algorithmname}

\usepackage{enumitem}		

\@ifundefined{date}{}{\date{}}
\usepackage{acl2015}
\usepackage{times}
\usepackage{url}
\usepackage{latexsym}
\usepackage{relsize}
\usepackage{multirow}
\usepackage{hyperref}
\usepackage{scrextend}
\usepackage{tikz}
\usepackage{placeins}
\usepackage{appendix}

\newcommand{\ignore}[1]{}

\title{A Generative Word Embedding Model and its Low Rank Positive Semidefinite Solution}

\author{Shaohua Li$^1$, Jun Zhu$^2$, Chunyan Miao$^1$ \\
  $^1$Joint NTU-UBC Research Centre of Excellence in Active Living for the Elderly (LILY), \\
  Nanyang Technological University, Singapore \\
  $^2$Tsinghua University, P.R. China \\
  {\tt lish0018@ntu.edu.sg, dcszj@tsinghua.edu.cn, ascymiao@ntu.edu.sg} \\
}

\newcommand{\vast}{\bBigg@{3}}
\newcommand{\Vast}{\bBigg@{4}}

\DeclareMathOperator*{\argmax}{arg\,max}
\DeclareMathOperator*{\argmin}{arg\,min}

\DeclareMathOperator*{\T}{\scriptscriptstyle \top}
\newcommand{\script}{\scriptscriptstyle}

\def\smallunderbrace#1{\mathop{\vtop{\m@th\ialign{##\crcr
   $\hfil\displaystyle{#1}\hfil$\crcr
   \noalign{\kern3\p@\nointerlineskip}%
   \small \upbracefill\crcr\noalign{\kern3\p@}}}}\limits}

\newcommand\undermat[2]{%
  \makebox[0pt][l]{$\smash{\smallunderbrace{\phantom{%
    \begin{matrix}#2\end{matrix}}}_{\text{$#1$}}}$}#2}

\newcommand{\overbar}[1]{\mkern 1.5mu\overline{\mkern-1.5mu#1\mkern-1.5mu}\mkern 1.5mu}

\makeatletter
\renewcommand*\env@matrix[1][*\c@MaxMatrixCols c]{%
  \hskip -\arraycolsep
  \let\@ifnextchar\new@ifnextchar
  \array{#1}}
\makeatother

\makeatother

\begin{document}
\maketitle 
\begin{abstract}
Most existing word embedding methods can be categorized into Neural
Embedding Models and Matrix Factorization (MF)-based methods. However
some models are opaque to probabilistic interpretation, and MF-based
methods, typically solved using Singular Value Decomposition (SVD),
may incur loss of corpus information. In addition, it is desirable
to incorporate global latent factors, such as topics, sentiments or
writing styles, into the word embedding model. Since generative models
provide a principled way to incorporate latent factors, we propose
a generative word embedding model, which is easy to interpret, and
can serve as a basis of more sophisticated latent factor models. The
model inference reduces to a low rank weighted positive semidefinite
approximation problem. Its optimization is approached by eigendecomposition
on a submatrix, followed by online blockwise regression, which is
scalable and avoids the information loss in SVD. In experiments on
7 common benchmark datasets, our vectors are competitive to word2vec,
and better than other MF-based methods. 
\end{abstract}

\section{Introduction}

\belowdisplayskip=5pt plus 3pt minus 3pt  \abovedisplayskip=5pt plus 3pt minus 3pt 

The task of word embedding is to model the distribution of a word
and its context words using their corresponding vectors in a Euclidean
space. Then by doing regression on the relevant statistics derived
from a corpus, a set of vectors are recovered which best fit these
statistics. These vectors, commonly referred to as the \textit{embeddings},
capture semantic/syntactic regularities between the words.

The core of a word embedding method is the \emph{link function} that
connects the input --- the embeddings, with the output --- certain
corpus statistics. Based on the link function, the objective function
is developed. The reasonableness of the link function impacts the
quality of the obtained embeddings, and different link functions are
amenable to different optimization algorithms, with different scalability.
Based on the forms of the link function and the optimization techniques,
most methods can be divided into two classes: the traditional \textit{neural
embedding models}, and more recent \textit{low rank matrix factorization
methods}.

The neural embedding models use the\textbf{ softmax} link function
to model the conditional distribution of a word given its context
(or vice versa) as a function of the embeddings. The normalizer in
the softmax function brings intricacy to the optimization, which is
usually tackled by gradient-based methods. The pioneering work was
\cite{bengio}. Later \newcite{three} propose three different link
functions. However there are interaction matrices between the embeddings
in all these models, which complicate and slow down the training,
hindering them from being trained on huge corpora. \newcite{cbow}
and \newcite{word2vec} greatly simplify the conditional distribution,
where the two embeddings interact directly. They implemented the well-known
``word2vec'', which can be trained efficiently on huge corpora.
The obtained embeddings show excellent performance on various tasks.

Low-Rank Matrix Factorization (MF in short) methods include various
link functions and optimization methods. The link functions are usually
not softmax functions. MF methods aim to reconstruct certain corpus
statistics matrix by the product of two low rank factor matrices.
The objective is usually to minimize the reconstruction error, optionally
with other constraints. %
In this line of research, \newcite{ppmi} find that ``word2vec''
is essentially doing stochastic weighted factorization of the word-context
pointwise mutual information (PMI) matrix. They then factorize this
matrix directly as a new method. \newcite{glove} propose a bilinear
regression function of the conditional distribution, from which a
weighted MF problem on the bigram log-frequency matrix is formulated.
Gradient Descent is used to find the embeddings. Recently, based on
the intuition that words can be organized in semantic hierarchies,
\newcite{sparsecoding2} add hierarchical sparse regularizers to
the matrix reconstruction error. With similar techniques, \newcite{sparsecoding1}
reconstruct a set of pretrained embeddings using sparse vectors of
greater dimensionality. \newcite{dhillon2} apply Canonical Correlation
Analysis (CCA) to the word matrix and the context matrix, and use
the canonical correlation vectors between the two matrices as word
embeddings. \newcite{stratos1} and \newcite{stratos2} assume
a Brown language model, and prove that doing CCA on the bigram occurrences
is equivalent to finding a transformed solution of the language model.
\newcite{randomwalk} assume there is a hidden discourse vector
on a random walk, which determines the distribution of the current
word. The slowly evolving discourse vector puts a constraint on the
embeddings in a small text window. The maximum likelihood estimate
of the embeddings within this text window approximately reduces to
a squared norm objective.

There are two limitations in current word embedding methods. The first
limitation is, all MF-based methods map words and their context words
to two different sets of embeddings, and then employ Singular Value
Decomposition (SVD) to obtain a low rank approximation of the word-context
matrix $\boldsymbol{M}$. As SVD factorizes $\boldsymbol{M}^{\T}\boldsymbol{M}$,
some information in $\boldsymbol{M}$ is lost, and the learned embeddings
may not capture the most significant regularities in $\boldsymbol{M}$.
\ref{sec:svdtrap} gives a toy example on which SVD does not work
properly.

The second limitation is, a generative model for documents parametered
by embeddings is absent in recent development. Although \cite{stratos1,stratos2,randomwalk}
are based on generative processes, the generative processes are only
for deriving the local relationship between embeddings within a small
text window, leaving the likelihood of a document undefined. In addition,
the learning objectives of some models, e.g. \cite[Eq.1]{word2vec},
even have no clear probabilistic interpretation. A generative word
embedding model for documents is not only easier to interpret and
analyze, but more importantly, provides a basis upon which document-level
\textit{global }latent factors, such as document topics \cite{bigramtopicmodel},
sentiments \cite{senttopic}, writing styles \cite{twittertraditional},
can be incorporated in a principled manner, to better model the text
distribution and extract relevant information.

Based on the above considerations, we propose to unify the embeddings
of words and context words. Our link function factorizes into three
parts: the interaction of two embeddings capturing linear correlations
of two words, a residual capturing nonlinear or noisy correlations,
and the unigram priors. To reduce overfitting, we put Gaussian priors
on embeddings and residuals, and apply Jelinek-Mercer Smoothing to
bigrams. Furthermore, to model the probability of a sequence of words,
we assume that the contributions of more than one context word approximately
add up. Thereby a generative model of documents is constructed, parameterized
by embeddings and residuals. The learning objective is to maximize
the corpus likelihood, which reduces to a weighted low-rank positive
semidefinite (PSD) approximation problem of the PMI matrix. A Block
Coordinate Descent algorithm is adopted to find an approximate solution.
This algorithm is based on Eigendecomposition, which avoids information
loss in SVD, but brings challenges to scalability. We then exploit
the sparsity of the weight matrix and implement an efficient online
blockwise regression algorithm. On seven benchmark datasets covering
similarity and analogy tasks, our method achieves competitive and
stable performance.

The source code of this method is provided at \begingroup\small\url{https://github.com/askerlee/topicvec}\endgroup.

\section{Notations and Definitions}

Throughout the paper, we always use a uppercase bold letter as $\boldsymbol{S},\boldsymbol{V}$
to denote a matrix or set, a lowercase bold letter as $\boldsymbol{v}_{w_{i}}$
to denote a vector, a normal uppercase letter as $N,W$ to denote
a scalar constant, and a normal lowercase letter as $s_{i},w_{i}$
to denote a scalar variable.

Suppose a vocabulary $\boldsymbol{S}=\{s_{1},\cdots,s_{\script W}\}$
consists of all the words, where $W$ is the vocabulary size. We further
suppose $s_{1},\cdots,s_{W}$ are sorted in decending order of the
frequency, i.e. $s_{1}$ is most frequent, and $s_{W}$ is least frequent.
A document $d_{i}$ is a sequence of words $d_{i}=(w_{i1},\cdots,w_{iL_{i}}),w_{ij}\in\boldsymbol{S}$.
A corpus is a collection of $M$ documents $\boldsymbol{D}=\{d_{1},\cdots,d_{\script M}\}$.
In the vocabulary, each word $s_{i}$ is mapped to a vector $\boldsymbol{v}_{s_{i}}$
in $N$-dimensional Euclidean space.

In a document, a sequence of words is referred to as a \textit{text
window}, denoted by $w_{i},\cdots,w_{i+l}$, or $w_{i}{:}w_{i+l}$
in shorthand. A text window of chosen size $c$ before a word $w_{i}$
defines the \textit{context} of $w_{i}$ as $w_{i-c},\cdots,w_{i-1}$.
Here $w_{i}$ is referred to as the \textit{focus word}. Each context
word $w_{i-j}$ and the focus word $w_{i}$ comprise a\textit{ }\textit{\emph{bigram}}
$w_{i-j},w_{i}$.

The \textit{Pointwise Mutual Information} between two words $s_{i},s_{j}$
is defined as\vspace{-0.15in}

\[
\textrm{PMI}(s_{i},s_{j})=\log\frac{P(s_{i},s_{j})}{P(s_{i})P(s_{j})}.
\]

\begin{table}
\centering{}%
\begin{tabular}{cc}
\hline 
Name  & Description\tabularnewline
\hline 
{\footnotesize{}$\boldsymbol{S}$ }  & {\footnotesize{}Vocabulary $\{s_{1},\cdots,s_{\script W}\}$}\tabularnewline
{\footnotesize{}$\boldsymbol{V}$}  & \textit{\emph{\footnotesize{}Embedding matrix $(\boldsymbol{v}_{s_{1}},\cdots,\boldsymbol{v}_{s_{\script W}})$}}\tabularnewline
{\footnotesize{}$\boldsymbol{D}$}  & {\footnotesize{}Corpus $\{d_{1},\cdots,d_{\script M}\}$}\tabularnewline
{\footnotesize{}$\boldsymbol{v}_{s_{i}}$ }  & \textit{\emph{\footnotesize{}Embedding }}{\footnotesize{}of word $s_{i}$}\tabularnewline
{\footnotesize{}$a_{s_{i}s_{j}}$ }  & \textit{\emph{\footnotesize{}Bigram residual}}{\footnotesize{} for
$s_{i},s_{j}$}\tabularnewline
{\footnotesize{}\negthickspace{}$\tilde{P}(s_{i},\negthinspace s_{j})$}  & {\footnotesize{}Empirical probability of $s_{i},s_{j}$ in the corpus}\tabularnewline
{\footnotesize{}$\boldsymbol{u}$ }  & {\footnotesize{}Unigram probability vector $\left(P(s_{1}),\negmedspace\cdots\negmedspace,P(s_{\script W})\right)$}\tabularnewline
{\footnotesize{}$\boldsymbol{A}$}  & {\footnotesize{}Residual matrix $(a_{s_{i}s_{j}})$}\tabularnewline
{\footnotesize{}$\boldsymbol{B}$}  & {\footnotesize{}Conditional probability matrix $\Bigl(P(s_{j}|s_{i})\Bigr)$}\tabularnewline
\multirow{1}{*}{{\footnotesize{}$\boldsymbol{G}$}}  & \textit{\emph{\footnotesize{}PMI matrix}}{\footnotesize{} $\Bigl(\textrm{PMI}(s_{i},s_{j})\Bigr)$}\tabularnewline
{\footnotesize{}$\boldsymbol{H}$}  & {\footnotesize{}Bigram empirical probability matrix $\Bigl(\tilde{P}(s_{i},s_{j})\Bigr)$}\tabularnewline
\hline 
\end{tabular}\protect\caption{Notation Table}
\end{table}

\section{Link Function of Text}

In this section, we formulate the probability of a sequence of words
as a function of their embeddings. We start from the link function
of bigrams, which is the building blocks of a long sequence. Then
this link function is extended to a text window with $c$ context
words, as a first-order approximation of the actual probability.

\subsection{Link Function of Bigrams\label{sub:bigram-distributions}}

We generalize the link function of ``word2vec'' and ``GloVe''
to the following: 
\begin{equation}
\hspace*{-0.9em}P(s_{i},s_{j})=\exp\left\{ \boldsymbol{v}_{s_{j}}^{\T}\boldsymbol{v}_{s_{i}}+a_{s_{i}s_{j}}\right\} \negmedspace P(s_{i})P(s_{j})\label{eq:embed2joint}
\end{equation}

The rationale for \eqref{eq:embed2joint} originates from the idea
of the \textit{Product of Experts} in \cite{productsexpert}. Suppose
different types of semantic/syntactic regularities between $s_{i}$
and $s_{j}$ are encoded in different dimensions of $\boldsymbol{v}_{s_{i}},\boldsymbol{v}_{s_{j}}$.
As $\exp\{\boldsymbol{v}_{s_{j}}^{\T}\boldsymbol{v}_{s_{i}}\}=\prod_{l}\exp\{v_{s_{i},l}\cdot v_{s_{j},l}\}$,
this means the effects of different regularities on the probability
are combined by multiplying together. If $s_{i}$ and $s_{j}$ are
independent, their joint probability should be $P(s_{i})P(s_{j})$.
In the presence of correlations, the actual joint probability $P(s_{i},s_{j})$
would be a scaling of it. The scale factor reflects how much $s_{i}$
and $s_{j}$ are positively or negatively correlated. Within the scale
factor, $\boldsymbol{v}_{s_{j}}^{\T}\boldsymbol{v}_{s_{i}}$ captures
linear interactions between $s_{i}$ and $s_{j}$, the residual $a_{s_{i}s_{j}}$
captures nonlinear or noisy interactions. In applications, only $\boldsymbol{v}_{s_{j}}^{\T}\boldsymbol{v}_{s_{i}}$
is of interest. Hence the bigger magnitude $\boldsymbol{v}_{s_{j}}^{\T}\boldsymbol{v}_{s_{i}}$
is of relative to $a_{s_{i}s_{j}}$, the better.

Note that we do \textit{not} assume $a_{s_{i}s_{j}}=a_{s_{j}s_{i}}$.
This provides the flexibility $P(s_{i},s_{j})\ne P(s_{j},s_{i})$,
agreeing with the asymmetry of bigrams in natural languages. At the
same time, $\boldsymbol{v}_{s_{j}}^{\T}\boldsymbol{v}_{s_{i}}$ imposes
a symmetric part between $P(s_{i},s_{j})$ and $P(s_{j},s_{i})$.

\eqref{eq:embed2joint} is equivalent to 
\begin{gather}
\hspace*{-0.9em}P(s_{j}|s_{i})\negmedspace=\negmedspace\exp\left\{ \boldsymbol{v}_{s_{j}}^{\T}\boldsymbol{v}_{s_{i}}+a_{s_{i}s_{j}}\negmedspace+\log P(s_{j})\right\} ,\label{eq:embed2cond}\\
\log\frac{P(s_{j}|s_{i})}{P(s_{j})}=\boldsymbol{v}_{s_{j}}^{\T}\boldsymbol{v}_{s_{i}}+a_{s_{i}s_{j}}.\label{eq:pmi2}
\end{gather}

\eqref{eq:pmi2} of all bigrams is represented in matrix form: 
\begin{equation}
\boldsymbol{V}^{\T}\boldsymbol{V}+\boldsymbol{A}=\boldsymbol{G},\label{eq:VV+A}
\end{equation}
where $\boldsymbol{G}$ is the PMI matrix.

\subsubsection{Gaussian Priors on Embeddings}

When \eqref{eq:embed2joint} is employed on the regression of empirical
bigram probabilities, a practical issue arises: more and more bigrams
have zero frequency as the constituting words become less frequent.
A zero-frequency bigram does not necessarily imply negative correlation
between the two constituting words; it could simply result from missing
data. But in this case, even after smoothing, \eqref{eq:embed2joint}
will force $\boldsymbol{v}_{s_{j}}^{\T}\boldsymbol{v}_{s_{i}}+a_{s_{i}s_{j}}$
to be a big negative number, making $\boldsymbol{v}_{s_{i}}$ overly
long. The increased magnitude of embeddings is a sign of overfitting.

To reduce overfitting of embeddings of infrequent words, we assign
a Spherical Gaussian prior $\mathcal{N}(0,\frac{1}{2\mu_{i}}\boldsymbol{I})$
to $\boldsymbol{v}_{s_{i}}$: 
\[
P(\boldsymbol{v}_{s_{i}})\sim\exp\{-\mu_{i}\Vert\boldsymbol{v}_{s_{i}}\Vert^{2}\},
\]
where the hyperparameter $\mu_{i}$ increases as the frequency of
$s_{i}$ decreases.

\subsubsection{Gaussian Priors on Residuals}

We wish $\boldsymbol{v}_{s_{j}}^{\T}\boldsymbol{v}_{s_{i}}$ in \eqref{eq:embed2joint}
captures as much correlations between $s_{i}$ and $s_{j}$ as possible.
Thus the smaller $a_{s_{i}s_{j}}$ is, the better. In addition, the
more frequent $s_{i},s_{j}$ is in the corpus, the less noise there
is in their empirical distribution, and thus the residual $a_{s_{i}s_{j}}$
should be more heavily penalized.

To this end, we penalize the residual $a_{s_{i}s_{j}}$ by $f(\tilde{{\scriptstyle P}}(s_{i},s_{j}))a_{s_{i}s_{j}}^{2}$,
where $f(\cdot)$ is a nonnegative monotonic transformation, referred
to as the \textit{weighting function}. Let $h_{ij}$ denote $\tilde{P}(s_{i},s_{j})$,
then the total penalty of all residuals are the square of the \textit{weighted}
\textit{Frobenius norm} of $\boldsymbol{A}$: 
\begin{equation}
\sum_{s_{i},s_{j}\in\boldsymbol{S}}\:f(h_{ij})a_{s_{i}s_{j}}^{2}=\Vert\boldsymbol{A}\Vert_{f(\boldsymbol{H})}^{2}.
\end{equation}

By referring to ``GloVe'', we use the following weighting function,
and find it performs well: 
\[
f(h_{ij})=\begin{cases}
\frac{\sqrt{h_{ij}}}{C_{\textrm{cut}}} & \sqrt{h_{ij}}<C_{\textrm{cut}},i\ne j\\
1 & \sqrt{h_{ij}}\ge C_{\textrm{cut}},i\ne j\\
0 & i=j
\end{cases},
\]
where $C_{\textrm{cut}}$ is chosen to cut the most frequent $0.02\%$
of the bigrams off at $1$. When $s_{i}=s_{j}$, two identical words
usually have much smaller probability to collocate. Hence $\tilde{P}(s_{i},s_{i})$
does not reflect the true correlation of a word to itself, and should
not put constraints to the embeddings. We eliminate their effects
by setting $f(h_{ii})$ to $0$.

If the domain of $\boldsymbol{A}$ is the whole space $R^{W\times W}$,
then this penalty is equivalent to a Gaussian prior $\mathcal{N}\left(0,\frac{1}{2f(h_{ij})}\right)$
on each $a_{s_{i}s_{j}}$. The variances of the Gaussians are determined
by the bigram empirical probability matrix $\boldsymbol{H}$.

\subsubsection{Jelinek-Mercer Smoothing of Bigrams}

As another measure to reduce the impact of missing data, we apply
the commonly used Jelinek-Mercer Smoothing \cite{smoothing} to smooth
the empirical conditional probability $\tilde{P}(s_{j}|s_{i})$ by
the unigram probability $\tilde{P}(s_{j})$ as: 
\begin{equation}
\tilde{P}_{\textrm{smoothed}}(s_{j}|s_{i})=(1-\kappa)\tilde{P}(s_{j}|s_{i})+\kappa P(s_{j}).\label{eq:smoothed_cond}
\end{equation}

Accordingly, the smoothed bigram empirical joint probability is defined
as 
\begin{equation}
\tilde{P}(s_{i},s_{j})=(1-\kappa)\tilde{P}(s_{i},s_{j})+\kappa P(s_{i})P(s_{j}).\label{eq:smoothed_joint}
\end{equation}

In practice, we find $\kappa=0.02$ yields good results. When $\kappa\ge0.04$,
the obtained embeddings begin to degrade with $\kappa$, indicating
that smoothing distorts the true bigram distributions.

\subsection{Link Function of a Text Window}

In the previous subsection, a regression link function of bigram probabilities
is established. In this section, we adopt a first-order approximation
based on Information Theory, and extend the link function to a longer
sequence $w_{0},\cdots,w_{c-1},w_{c}$.

Decomposing a distribution conditioned on $n$ random variables as
the conditional distributions on its subsets roots deeply in Information
Theory. This is an intricate problem because there could be both \textit{\emph{(pointwise)}}\textit{
redundant information} and \textit{\emph{(pointwise)}}\textit{ synergistic
information} among the conditioning variables \cite{multinfo}. They
are both functions of the PMI. Based on an analysis of the complementing
roles of these two types of pointwise information, we assume they
are approximately equal and cancel each other when computing the \emph{pointwise
interaction information}. See Appendix B for a detailed discussion.

Following the above assumption, we have $\textrm{PMI}(w_{2};w_{0},w_{1})\approx\textrm{PMI}(w_{2};w_{0})+\textrm{PMI}(w_{2};w_{1})$:
\[
\log\negthickspace\frac{{\scriptstyle P}(w_{0},w_{1}|w_{2})}{{\scriptstyle P}(w_{0},w_{1})}\negmedspace\approx\negmedspace\log\negthickspace\frac{{\scriptstyle P}(w_{0}|w_{2})}{{\scriptstyle P}(w_{0})}\negmedspace+\negmedspace\log\negthickspace\frac{{\scriptstyle P}(w_{1}|w_{2})}{{\scriptstyle P}(w_{1})}.
\]

Plugging \eqref{eq:embed2joint} and \eqref{eq:pmi2} into the above,
we obtain 
\begin{align*}
 & P(w_{0},w_{1},w_{2})\\
\approx & \exp\biggl\{\sum_{\substack{i,j=0\\
i\ne j
}
}^{2}(\boldsymbol{v}_{w_{i}}^{\T}\boldsymbol{v}_{w_{j}}+a_{w_{i}w_{j}})+\sum_{i=0}^{2}\log P(w_{i})\biggr\}.
\end{align*}

We extend the above assumption to that the pointwise interaction information
is still close to $0$ within a longer text window. Accordingly the
above equation extends to a context of size $c>2$: 
\begin{align*}
 & P(w_{0},\cdots,w_{c})\\
\approx & \exp\biggl\{\sum_{\substack{i,j=0\\
i\ne j
}
}^{c}(\boldsymbol{v}_{w_{i}}^{\T}\boldsymbol{v}_{w_{j}}+a_{w_{i}w_{j}})+\sum_{i=0}^{c}\log P(w_{i})\biggr\}.
\end{align*}

From it derives the conditional distribution of $w_{c}$, given its
context $w_{0},\cdots,w_{c-1}$: 
\begin{align}
 & P(w_{c}\mid w_{0}:w_{c-1})\negmedspace=\negmedspace\frac{P(w_{0},\cdots,w_{c})}{P(w_{0},\cdots,w_{c-1})}\nonumber \\
\approx & P(w_{c})\exp\biggl\{\boldsymbol{v}_{w_{c}}^{\T}\sum_{i=0}^{c-1}\boldsymbol{v}_{w_{i}}+\sum_{i=0}^{c-1}a_{w_{i}w_{c}}\biggr\}.\label{eq:groupcond}
\end{align}

\section{Generative Process and Likelihood}

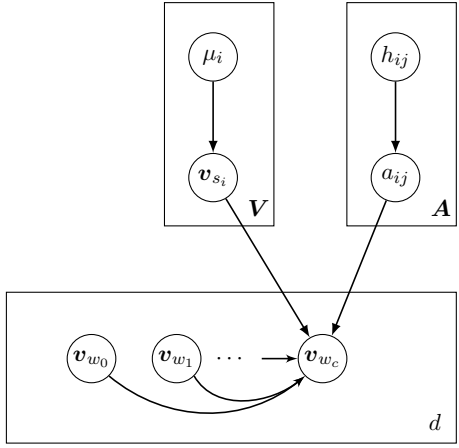
\begin{figure}
\centering{}\usetikzlibrary{arrows} \usetikzlibrary{decorations.markings}
\begin{tikzpicture}[scale=0.8, every node/.style={transform shape, inner sep=-4pt, minimum width=0.8cm}]
\node [circle, draw] (w0) at (1,4) {$\boldsymbol{v}_{w_0}$}; \node (w1) at (2.4,4) [circle, draw] {$\boldsymbol{v}_{w_1}$}; \node (wc) at (4.8,4) [circle, draw] {$\boldsymbol{v}_{w_c}$};
\node (dots2) at (3.3,4) [draw=none] {$\cdots$};
\draw[>=latex',->,semithick] (w0.south east) to[out=-40,in=220] (wc.south west); \draw[>=latex',->,semithick] (w1.south east) to[out=-60,in=220] (wc.south west); \draw[>=latex',->,semithick] (3.8,4) -- (wc.west);
\draw (-0.4,2.6) rectangle (7,5.1);
\node at (6.65,2.95) {$d$}; \draw  (2.2,9.9) rectangle (4,6.2); \draw  (5.2,9.9) rectangle (7,6.2); \node at (3.75,6.5) {$\boldsymbol{V}$}; \node at (6.75,6.5) {$\boldsymbol{A}$}; \node [circle,draw] (mu_i) at (3,9) {$\mu_i$}; \node [circle,draw] (vsi) at (3,7) {$\boldsymbol{v}_{s_i}$}; \node [circle,draw] (hij) at (6,9) {$h_{ij}$}; \node [circle,draw] (aij) at (6,7) {$a_{ij}$}; \draw[>=latex,->,semithick]   (mu_i) -- (vsi); \draw[>=latex,->,semithick]   (hij) -- (aij); \draw[>=latex,->,semithick]   (vsi) -- (wc); \draw[>=latex,->,semithick]   (aij) -- (wc);
\end{tikzpicture} \protect\caption{\label{fig:GM}The Graphical Model of PSDVec}
\end{figure}

We proceed to assume the text is generated from a \textit{Markov chain}
of order $c$, i.e., a word only depends on words within its context
of size $c$. Given the hyperparameter $\boldsymbol{\mu}=(\mu_{1},\cdots\negmedspace,\mu_{\script W})$,
the generative process of the whole corpus is:
\begin{enumerate}[topsep=4pt,itemsep=-0.6ex]
\item For each word $s_{i}$, draw the embedding $\boldsymbol{v}_{s_{i}}$
from\, $\mathcal{N}(0,\frac{1}{2\mu_{i}}\boldsymbol{I})$;
\item For each bigram $s_{i},s_{j}$, draw the residual $a_{s_{i}s_{j}}$
from\, $\mathcal{N}\left(0,\frac{1}{2f(h_{ij})}\right)$;
\item For each document $d_{i}$, for the $j$-th word, draw word $w_{ij}$
from $\boldsymbol{S}$ with probability $P(w_{ij}\mid w_{i,j-c}:w_{i,j-1})$
defined by \eqref{eq:groupcond}. 
\end{enumerate}
The above generative process for a document $d$ is presented as a
graphical model in Figure \ref{fig:GM}.

Based on this generative process, the probability of a document $d_{i}$
can be derived as follows, given the embeddings and residuals $\boldsymbol{V},\boldsymbol{A}$:
\begin{align*}
 & P(d_{i}|\boldsymbol{V},\boldsymbol{A})\\
= & \prod_{j=1}^{L_{i}}P(w_{ij})\exp\biggl\{\boldsymbol{v}_{w_{ij}}^{\T}\negmedspace\sum_{k=j-c}^{j-1}\boldsymbol{v}_{w_{ik}}\negmedspace+\negmedspace\sum_{k=j-c}^{j-1}\negmedspace a_{w_{ik}w_{ij}}\biggr\}.
\end{align*}

The complete-data likelihood of the corpus is:\vspace{-0.25in}\begin{addmargin}[-0.5em]{0em}
\begin{align*}
 & p(\boldsymbol{D},\boldsymbol{V},\boldsymbol{A})\\
= & \prod_{i=1}^{W}\mathcal{N}(0,\frac{\boldsymbol{I}}{2\mu_{i}})\prod_{i,j=1}^{W,W}\negmedspace\mathcal{N}\left(0,\frac{1}{2f(h_{ij})}\right)\prod_{i=1}^{M}p(d_{i}|\boldsymbol{V}\negmedspace,\boldsymbol{A})\\
= & \frac{1}{\mathcal{Z}(\boldsymbol{H},\boldsymbol{\mu})}\exp\Bigl\{-\negmedspace\sum_{i,j=1}^{W,W}f(h_{i,j})a_{s_{i}s_{j}}^{2}\negmedspace-\negmedspace\sum_{i=1}^{W}\mu_{i}\Vert\boldsymbol{v}_{s_{i}}\Vert^{2}\Bigr\}\\
 & \cdot\negthickspace\prod_{i,j=1}^{M,L_{i}}\negmedspace P(w_{ij})\exp\biggl\{\boldsymbol{v}_{w_{ij}}^{\T}\negmedspace\sum_{k=j-c}^{j-1}\negmedspace\boldsymbol{v}_{w_{ik}}\negmedspace+\negmedspace\negmedspace\sum_{k=j-c}^{j-1}\negmedspace a_{w_{ik}w_{ij}}\biggr\},
\end{align*}
\end{addmargin} where $\mathcal{Z}(\boldsymbol{H},\boldsymbol{\mu})$
is the normalizing constant.

Taking the logarithm of both sides of $p(\boldsymbol{D},\boldsymbol{A},\boldsymbol{V})$
yields 
\begin{align}
 & \log p(\boldsymbol{D},\boldsymbol{V},\boldsymbol{A})\nonumber \\
= & C_{0}-\log\mathcal{Z}(\boldsymbol{H},\boldsymbol{\mu})-\Vert\boldsymbol{A}\Vert_{f(\boldsymbol{H})}^{2}\negmedspace-\negmedspace\sum_{i=1}^{W}\mu_{i}\Vert\boldsymbol{v}_{s_{i}}\Vert^{2}\nonumber \\
 & \negmedspace+\negmedspace\sum_{i,j=1}^{M,L_{i}}\biggl\{\boldsymbol{v}_{w_{ij}}^{\T}\negmedspace\sum_{k=j-c}^{j-1}\boldsymbol{v}_{w_{ik}}\negmedspace+\negmedspace\sum_{k=j-c}^{j-1}a_{w_{ik}w_{ij}}\biggr\},\label{eq:logjointcorpus}
\end{align}
where $C_{0}=\sum_{i,j=1}^{M,L_{i}}\log P(w_{ij})$ is constant.\medskip{}

\section{Learning Algorithm}

\subsection{Learning Objective}

The learning objective is to find the embeddings $\boldsymbol{V}$
that maximize the corpus log-likelihood \eqref{eq:logjointcorpus}.

Let $x_{ij}$ denote the (smoothed) frequency of bigram $s_{i},s_{j}$
in the corpus. Then \eqref{eq:logjointcorpus} is sorted as: 
\begin{align}
 & \log p(\boldsymbol{D},\boldsymbol{V},\boldsymbol{A})\nonumber \\
= & C_{0}-\log\mathcal{Z}(\boldsymbol{H},\boldsymbol{\mu})-\Vert\boldsymbol{A}\Vert_{f(\boldsymbol{H})}^{2}-\sum_{i=1}^{W}\mu_{i}\Vert\boldsymbol{v}_{s_{i}}\Vert^{2}\nonumber \\
 & +\sum_{i,j=1}^{W,W}x_{ij}(\boldsymbol{v}_{s_{i}}^{\T}\boldsymbol{v}_{s_{j}}+a_{s_{i}s_{j}}).\label{eq:logjointcorpus2}
\end{align}

As the corpus size increases, $\sum_{i,j=1}^{W,W}x_{ij}(\boldsymbol{v}_{s_{i}}^{\T}\boldsymbol{v}_{s_{j}}\negmedspace+\negmedspace a_{s_{i}s_{j}})$
will dominate the parameter prior terms. Then we can ignore the prior
terms when maximizing \eqref{eq:logjointcorpus2}. 
\begin{align*}
 & \max\sum x_{ij}(\boldsymbol{v}_{s_{i}}^{\T}\boldsymbol{v}_{s_{j}}\negmedspace+\negmedspace a_{s_{i}s_{j}})\\
= & \left(\sum x_{ij}\right)\cdot\max\sum\tilde{P}_{\textrm{smoothed}}(s_{i},s_{j})\log P(s_{i},s_{j}).
\end{align*}

As both $\{\tilde{P}_{\textrm{smoothed}}(s_{i},s_{j})\}$ and $\{P(s_{i},s_{j})\}$
sum to 1, the above sum is maximized when $P(s_{i},s_{j})=\tilde{P}_{\textrm{smoothed}}(s_{i},s_{j}).$

The maximum likelihood estimator is then: 
\begin{align}
P(s_{j}|s_{i}) & =\tilde{P}_{\textrm{smoothed}}(s_{j}|s_{i}),\nonumber \\
\boldsymbol{v}_{s_{i}}^{\T}\boldsymbol{v}_{s_{j}}+a_{s_{i}s_{j}} & =\log\frac{\tilde{P}_{\textrm{smoothed}}(s_{j}|s_{i})}{P(s_{j})}.\label{eq:p_v_sol}
\end{align}

Writing \eqref{eq:p_v_sol} in matrix form: 
\begin{align}
\boldsymbol{B}^{*} & =\Bigl(\tilde{P}_{\textrm{smoothed}}(s_{j}|s_{i})\Bigr)_{s_{i},s_{j}\in\boldsymbol{S}}\nonumber \\
\boldsymbol{G}^{*} & =\log\boldsymbol{B}^{*}-\log\boldsymbol{u}\otimes(1\cdots1),\label{eq:B_G_est}
\end{align}
where ``$\otimes$'' is the outer product. \vspace{3pt}

Now we fix the values of $\boldsymbol{v}_{s_{i}}^{\T}\boldsymbol{v}_{s_{j}}+a_{s_{i}s_{j}}$
at the above optimal. The corpus likelihood becomes
\begin{align}
\log p(\boldsymbol{D},\boldsymbol{V},\boldsymbol{A})= & C_{1}-\Vert\boldsymbol{A}\Vert_{f(\boldsymbol{H})}^{2}-\sum_{i=1}^{W}\mu_{i}\Vert\boldsymbol{v}_{s_{i}}\Vert^{2},\nonumber \\
\textrm{subject to\enskip\  } & \hspace{0.2em}\boldsymbol{V}^{\T}\boldsymbol{V}+\boldsymbol{A}=\boldsymbol{G}^{*},\label{eq:logjointcorpus3}
\end{align}
where $C_{1}=C_{0}+\sum x_{ij}\log\tilde{P}_{\textrm{smoothed}}(s_{i},s_{j})-\log\mathcal{Z}(\boldsymbol{H},\boldsymbol{\mu})$
is constant.\medskip{}

\subsection{\texorpdfstring{Learning $\boldsymbol{V}$ as Low Rank PSD Approximation}{Learning
V as Low Rank PSD Approximation}}

Once $\boldsymbol{G}^{*}$ has been estimated from the corpus using
\eqref{eq:B_G_est}, we seek $\boldsymbol{V}$ that maximizes \eqref{eq:logjointcorpus3}.
This is to find the maximum a posteriori (MAP) estimates of $\boldsymbol{V},\boldsymbol{A}$
that satisfy $\boldsymbol{V}^{\T}\boldsymbol{V}+\boldsymbol{A}=\boldsymbol{G}^{*}$.
Applying this constraint to \eqref{eq:logjointcorpus3}, we obtain\vspace{-0.2in}\begin{addmargin}[-1em]{0em}\begingroup\addtolength{\jot}{-5pt}
\begin{flalign}
 & \argmax_{\boldsymbol{V}}\,\log p(\boldsymbol{D},\boldsymbol{V},\boldsymbol{A})\nonumber \\
= & \argmin_{\boldsymbol{V}}\Vert\boldsymbol{G}^{*}\negmedspace-\negmedspace\boldsymbol{V}^{\T}\boldsymbol{V}\Vert_{f(\boldsymbol{H})}+\negmedspace\sum_{i=1}^{W}\mu_{i}\Vert\boldsymbol{v}_{s_{i}}\Vert^{2}.\label{eq:minG-VV}
\end{flalign}
\endgroup\end{addmargin}

Let $\boldsymbol{X}=\boldsymbol{V}^{\T}\boldsymbol{V}$. Then $\boldsymbol{X}$
is positive semidefinite of rank $N$. Finding $\boldsymbol{V}$ that
minimizes \eqref{eq:minG-VV} is equivalent to finding a rank-$N$
\textit{weighted positive semidefinite} \textit{approximant }$\boldsymbol{X}$
of \textit{$\boldsymbol{G}^{*}$}, subject to Tikhonov regularization.
This problem does not admit an analytic solution, and can only be
solved using local optimization methods.

First we consider a simpler case where all the words in the vocabulary
are enough frequent, and thus Tikhonov regularization is unnecessary.
In this case, we set $\forall\mu_{i}=0$, and \eqref{eq:minG-VV}
becomes an unregularized optimization problem. We adopt the Block
Coordinate Descent (BCD) algorithm\footnote{It is referred to as an Expectation-Maximization algorithm by the
original authors, but we think this is a misnomer.} in \cite{weightedlow} to approach this problem. The original algorithm
is to find a generic rank-$N$ matrix for a weighted approximation
problem, and we tailor it by constraining the matrix within the positive
semidefinite manifold\textit{.}

\begin{algorithm}
\protect\caption{\label{alg:BCD}BCD algorithm for finding a unregularized rank-$N$
weighted PSD approximant.}

\textbf{Input}: matrix $\boldsymbol{G}^{*}$, weight matrix $\boldsymbol{W}=f(\boldsymbol{H})$,

iteration number $\mathcal{T}$, rank $N$

\vspace{6pt}

Randomly initialize $\boldsymbol{X}^{(0)}$

\textbf{for} $t=1,\cdots,\mathcal{T}$ \textbf{do}

\textbf{\quad{}}$\boldsymbol{G}_{t}=\boldsymbol{W}\circ\boldsymbol{G}^{*}+(1-\boldsymbol{W})\circ\boldsymbol{X}^{(t-1)}$

\textbf{\quad{}}$\boldsymbol{X}^{(t)}=\textrm{PSD\_Approximate(}\boldsymbol{G}_{t},N)$

\textbf{end for}\vspace{2pt}

$\boldsymbol{\lambda},\boldsymbol{Q}$ = Eigen\_Decomposition($\boldsymbol{X}^{(\mathcal{T})}$)

$\boldsymbol{V}^{*}$ = $\textrm{diag}(\boldsymbol{\lambda}^{\frac{1}{2}}[1{:}N])\cdot\boldsymbol{Q}^{\T}[1{:}N]$\vspace{6pt}

\textbf{Output}: $\boldsymbol{V}^{*}$ 
\end{algorithm}

We summarize our learning algorithm in \textbf{Algorithm \ref{alg:BCD}.}
Here ``$\circ$'' is the entry-wise product. We suppose the eigenvalues
$\boldsymbol{\lambda}$ returned by Eigen\_Decomposition($\boldsymbol{X}$)
are in descending order. $\boldsymbol{Q}^{\T}[1{:}N]$ extracts the
1 to $N$ rows from $\boldsymbol{Q}^{\T}$.

One key issue is how to initialize $\boldsymbol{X}$. \newcite{weightedlow}
suggest to set $\boldsymbol{X}^{(0)}\negmedspace=\negmedspace\boldsymbol{G}^{*}$,
and point out that $\boldsymbol{X}^{(0)}\negmedspace=\negmedspace\boldsymbol{0}$
is far from a local optimum, thus requires more iterations. However
we find $\boldsymbol{G}^{*}$ is also far from a local optimum, and
this setting converges slowly too. Setting $\boldsymbol{X}^{(0)}=\boldsymbol{G}^{*}/2$
usually yields a satisfactory solution in a few iterations.

The subroutine $\textrm{PSD\_Approximate(})$ computes the \emph{unweighted}
nearest rank-$N$ PSD approximation, measured in F-norm \cite{nearPSD}.

\subsection{Online Blockwise Regression of \texorpdfstring{$\boldsymbol{V}$}{V}}

In Algorithm \ref{alg:BCD}, the essential subroutine PSD\_Approximate()
does eigendecomposition on $\boldsymbol{G}_{t}$, which is dense due
to the logarithm transformation. Eigendecomposition on a $W\times W$
dense matrix requires $O(W^{2})$ space and $O(W^{3})$ time, difficult
to scale up to a large vocabulary. In addition, the majority of words
in the vocabulary are infrequent, and Tikhonov regularization is necessary
for them.

It is observed that, as words become less frequent, fewer and fewer
words appear around them to form bigrams. Remind that the vocabulary
$\boldsymbol{S}=\{s_{1},\cdots,s_{W}\}$ are sorted in decending order
of the frequency, hence the lower-right blocks of $\boldsymbol{H}$
and $f(\boldsymbol{H})$ are very sparse, and cause these blocks in
\eqref{eq:minG-VV} to contribute much less penalty relative to other
regions. Therefore these blocks could be ignored when doing regression,
without sacrificing too much accuracy. This intuition leads to the
following \emph{online blockwise regression}.

The basic idea is to select a small set (e.g. 30,000) of the most
frequent words as the \emph{core words}, and partition the remaining
\emph{noncore words} into sets of moderate sizes. Bigrams consisting
of two core words are referred to as \emph{core bigrams}, which correspond
to the top-left blocks of $\boldsymbol{G}$ and $f(\boldsymbol{H})$.
The embeddings of core words are learned approximately using Algorithm
\ref{alg:BCD}, on the top-left blocks of $\boldsymbol{G}$ and $f(\boldsymbol{H})$.
Then we fix the embeddings of core words, and find the embeddings
of each set of noncore words in turn. After ignoring the lower-right
regions of $\boldsymbol{G}$ and $f(\boldsymbol{H})$ which correspond
to bigrams of two noncore words, the \textbf{quadratic terms} of noncore
embeddings are ignored. Consequently, finding these embeddings becomes
a \emph{weighted ridge regression} problem, which can be solved efficiently
in closed-form. Finally we combine all embeddings to get the embeddings
of the whole vocabulary. The details are as follows:
\begin{enumerate}[topsep=2pt,itemsep=-0.2ex]
\item Partition $\boldsymbol{S}$ into $K$ consecutive groups $\boldsymbol{S}_{1},\cdots,\boldsymbol{S}_{k}$.
Take $K=3$ as an example. The first group is core words;
\item Accordingly partition $\boldsymbol{G}$ into $K\times K$ blocks,
in this example as $\left(\begin{array}{c|cc}
\boldsymbol{G}_{11} & \boldsymbol{G}_{12} & \boldsymbol{G}_{13}\\
\hline \boldsymbol{G}_{21} & \boldsymbol{G}_{22} & \boldsymbol{G}_{23}\\
\boldsymbol{G}_{31} & \boldsymbol{G}_{32} & \boldsymbol{G}_{33}
\end{array}\right).$ Partition $f(\boldsymbol{H})$,$\boldsymbol{A}$ in the same way.
$\boldsymbol{G}_{11},f(\boldsymbol{H})_{11},\boldsymbol{A}_{11}$
correspond to core bigrams. Partition $\boldsymbol{V}$ into $\begin{array}{c}
\begin{pmatrix}[c|cc]\undermat{\boldsymbol{S}_{1}}{\boldsymbol{V}_{1}} & \negthickspace\undermat{\boldsymbol{S}_{2}}{\;\boldsymbol{V}_{2}} & \negthickspace\undermat{\boldsymbol{S}_{3}}{\;\boldsymbol{V}_{3}}\end{pmatrix}\\
\rule{0pt}{15pt}
\end{array}$;
\item Solve $\boldsymbol{V}_{1}^{\T}\boldsymbol{V}_{1}+\boldsymbol{A}_{11}=\boldsymbol{G}_{11}$
using Algorithm \ref{alg:BCD}, and obtain core embeddings $\boldsymbol{V}_{1}^{*}$;
\item Set $\boldsymbol{V}_{1}=\boldsymbol{V}_{1}^{*}$, and find $\boldsymbol{V}_{2}^{*}$
that minimizes the total penalty of the $12$-th and 21-th blocks
of residuals (the 22-th block is ignored due to its high sparsity):\vspace{-0.2in}\begin{addmargin}[-1em]{0em}\begingroup
\addtolength{\jot}{-2pt}
\begin{align*}
 & \argmin_{\boldsymbol{V}_{2}}\Vert\boldsymbol{G}_{12}-\boldsymbol{V}_{1}^{\T}\boldsymbol{V}_{2}\Vert_{f(\boldsymbol{H})_{12}}^{2}\\
 & \hspace{0.5em}+\Vert\boldsymbol{G}_{21}-\boldsymbol{V}_{2}^{\T}\boldsymbol{V}_{1}\Vert_{f(\boldsymbol{H})_{21}}^{2}+\sum_{s_{i}\in\boldsymbol{S}_{2}}\mu_{i}\Vert\boldsymbol{v}_{s_{i}}\Vert^{2}\\
= & \argmin_{\boldsymbol{V}_{2}}\Vert\overbar{\boldsymbol{G}}_{12}\negmedspace-\negmedspace\boldsymbol{V}_{1}^{\T}\boldsymbol{V}_{2}\Vert_{\bar{f}(\boldsymbol{H})_{12}}^{2}\negmedspace+\negmedspace\negmedspace\sum_{s_{i}\in\boldsymbol{S}_{2}}\negmedspace\negmedspace\mu_{i}\Vert\boldsymbol{v}_{s_{i}}\Vert^{2},
\end{align*}
\endgroup\end{addmargin} where $\bar{f}(\boldsymbol{H})_{12}=f(\boldsymbol{H})_{12}+f(\boldsymbol{H})_{21}^{\T}$;
$\overbar{\boldsymbol{G}}_{12}=\Bigl(\boldsymbol{G}_{12}\circ f(\boldsymbol{H})_{12}+\boldsymbol{G}_{21}^{\T}\circ f(\boldsymbol{H})_{21}^{\T}\Bigr)\allowbreak/\Bigl(f(\boldsymbol{H})_{12}+f(\boldsymbol{H})_{21}^{\T}\Bigr)$
is the weighted average of $\boldsymbol{G}_{12}$ and $\boldsymbol{G}_{21}^{\T}$,
``$\circ$'' and ``$/$'' are element-wise product and division,
respectively. The columns in $\boldsymbol{V}_{2}$ are independent,
thus for each $\boldsymbol{v}_{s_{i}}$, it is a separate weighted
ridge regression problem, whose solution is \cite{tikhonov}:
\[
\boldsymbol{v}_{s_{i}}^{*}\hspace{-0.3em}=\hspace{-0.3em}(\boldsymbol{V}_{1}^{\T}\textrm{diag}(\boldsymbol{\bar{f}}_{i})\boldsymbol{V}_{1}\negmedspace+\negmedspace\mu_{i}\boldsymbol{I})^{-1}\boldsymbol{V}_{1}^{\T}\textrm{diag}(\boldsymbol{\bar{f}}_{i})\bar{\boldsymbol{g}}_{i},
\]
where $\boldsymbol{\bar{f}}_{i}$ and $\bar{\boldsymbol{g}}_{i}$
are columns corresponding to $s_{i}$ in $\bar{f}(\boldsymbol{H})_{12}\,$
and $\overbar{\boldsymbol{G}}_{12}$, respectively;
\item For any other set of noncore words $\boldsymbol{S}_{k}$, find $\boldsymbol{V}_{k}^{*}$
that minimizes the total penalty of the $1k$-th and $k1$-th blocks,
ignoring all other $kj$-th and $jk$-th blocks; 
\item Combine all subsets of embeddings to form $\boldsymbol{V}^{*}$. Here
$\boldsymbol{V}^{*}=(\boldsymbol{V}_{1}^{*},\boldsymbol{V}_{2}^{*},\boldsymbol{V}_{3}^{*})$.
\end{enumerate}

\section{Experimental Results}

We trained our model along with a few state-of-the-art competitors
on Wikipedia, and evaluated the embeddings on 7 common benchmark sets.

\subsection{Experimental Setup}

Our own method is referred to as \textbf{PSD}. The competitors include: 
\begin{itemize}[topsep=4pt,itemsep=-0.6ex]
\item \cite{word2vec}: \textbf{word2vec}\footnote{https://code.google.com/p/word2vec/},
or SGNS in some literature; 
\item \cite{ppmi}: the \textbf{PPMI} matrix without dimension reduction,
and \textbf{SVD} of PPMI matrix, both yielded by hyperwords; 
\item \cite{glove}: \textbf{GloVe}\footnote{http://nlp.stanford.edu/projects/glove/}; 
\item \cite{stratos2}: \textbf{Singular}\footnote{https://github.com/karlstratos/singular},
which does SVD-based CCA on the weighted bigram frequency matrix; 
\item \cite{sparsecoding1}: \textbf{Sparse}\footnote{https://github.com/mfaruqui/sparse-coding},
which learns new sparse embeddings in a higher dimensional space from
pretrained embeddings. 
\end{itemize}
All models were trained on the English Wikipedia snapshot in March
2015. After removing non-textual elements and non-English words, 2.04
billion words were left. We used the default hyperparameters in Hyperwords
when training PPMI and SVD. Word2vec, GloVe and Singular were trained
with their own default hyperparameters. 

\begin{table*}[t]
\begin{centering}
\protect\protect\caption{\label{tab:scores}Performance of each method across different tasks.}
\vspace{5pt}
 
\par\end{centering}

\centering{}%
\begin{tabular}{|c|c|c|c|c|c|c|c|}
\hline 
 & \multicolumn{5}{c|}{Similarity Tasks} & \multicolumn{2}{c|}{Analogy Tasks}\tabularnewline
\hline 
Method  & WS Sim  & WS Rel  & MEN  & Turk  & SimLex  & Google  & MSR\tabularnewline
\hline 
\hline 
word2vec  & 0.742  & 0.543  & 0.731  & 0.663  & 0.395  & \textbf{0.734 }/ \textbf{0.742}  & \textbf{0.650} / \textbf{0.674}\tabularnewline
\hline 
PPMI  & 0.735  & 0.678  & 0.717  & 0.659  & 0.308  & 0.476 / 0.524  & 0.183 / 0.217\tabularnewline
\hline 
SVD  & 0.687  & 0.608  & 0.711  & 0.524  & 0.270  & 0.230 / 0.240  & 0.123 / 0.113\tabularnewline
\hline 
GloVe  & 0.759  & 0.630  & 0.756  & 0.641  & 0.362  & 0.535 / 0.544  & 0.408 / 0.435\tabularnewline
\hline 
Singular  & 0.763  & \textbf{0.684}  & 0.747  & 0.581  & 0.345  & 0.440 / 0.508  & 0.364 / 0.399\tabularnewline
\hline 
Sparse  & 0.739  & 0.585  & 0.725  & 0.625  & 0.355  & 0.240 / 0.282  & 0.253 / 0.274\tabularnewline
\hline 
PSD-Reg-180K  & \textbf{0.792}  & 0.679  & \textbf{0.764}  & \textbf{0.676}  & \textbf{0.398}  & 0.602 / 0.623  & 0.465 / 0.507\tabularnewline
\hline 
PSD-Unreg-180K & 0.786 & 0.663 & 0.753 & 0.675 & 0.372 & 0.566 / 0.598 & 0.424 / 0.468\tabularnewline
\hline 
PSD-25K  & \emph{0.801}  & \emph{0.676}  & \emph{0.765}  & \emph{0.678}  & \emph{0.393}  & \emph{0.671 / 0.695}  & \emph{0.533 / 0.586}\tabularnewline
\hline 
\end{tabular}
\end{table*}

The embedding sets PSD-Reg-180K and PSD-Unreg-180K were trained using
our online blockwise regression. Both sets contain the embeddings
of the most frequent 180,000 words, based on 25,000 core words. PSD-Unreg-180K
was traind with all $\mu_{i}=0$, i.e. disabling Tikhonov regularization.
PSD-Reg-180K was trained with $\mu_{i}=\begin{cases}
2 & i\in[25001,80000]\\
4 & i\in[80001,130000]\\
8 & i\in[130001,180000]
\end{cases},$ i.e. increased regularization as the sparsity increases. To contrast
with the batch learning performance, the performance of PSD-25K is
listed, which contains the core embeddings only. PSD-25K took advantages
that it contains much less false candidate words, and some test tuples
(generally harder ones) were not evaluated due to missing words, thus
its scores are not comparable to others.

Sparse was trained with PSD-180K-reg as the input embeddings, with
default hyperparameters.

The benchmark sets are almost identical to those in \cite{tacl},
except that \cite{luong}'s Rare Words is not included, as many rare
words are cut off at the frequency 100, making more than 1/3 of test
pairs invalid.

\textbf{Word Similarity} There are 5 datasets: WordSim Similarity
(\textbf{WS Sim}) and WordSim Relatedness (\textbf{WS Rel}) \cite{wordsim1,wordsim2},
partitioned from WordSim353 \cite{wordsim}; \newcite{bruni}'s
\textbf{MEN} dataset; \newcite{turk}'s Mechanical \textbf{Turk}
dataset; and \cite{simlex}'s \textbf{SimLex}-999 dataset. The embeddings
were evaluated by the Spearman's rank correlation with the human ratings.

\textbf{Word Analogy }The two datasets are \textbf{MSR}'s analogy
dataset \cite{msr}, containing 8000 questions, and \textbf{Google}'s
analogy dataset \cite{cbow}, with 19544 questions. After filtering
questions involving out-of-vocabulary words, i.e. words that appear
less than 100 times in the corpus, 7054 instances in MSR and 19364
instances in Google were left. The analogy questions were answered
using 3CosAdd as well as 3CosMul proposed by \newcite{3cos}.

\subsection{Results}

Table \ref{tab:scores} shows the results on all tasks. Word2vec significantly
outperformed other methods on analogy tasks. PPMI and SVD performed
much worse on analogy tasks than reported in \cite{tacl}, probably
due to sub-optimal hyperparameters. This suggests their performance
is unstable. \textit{\emph{The new embeddings yielded by Sparse systematically
}}\textit{degraded}\textit{\emph{ compared to the old embeddings,
contradicting the claim in \cite{sparsecoding1}.}}

Our method PSD-Reg-180K performed well consistently, and is best in
4 similarity tasks. It performed worse than word2vec on analogy tasks,
but still better than other MF-based methods. By comparing to PSD-Unreg-180K,
we see Tikhonov regularization brings 1-4\% performance boost across
tasks. In addition, on similarity tasks, online blockwise regression
only degrades slightly compared to batch factorization. Their performance
gaps on analogy tasks were wider, but this might be explained by the
fact that some hard cases were not counted in PSD-25K's evaluation,
due to its limited vocabulary.

\section{Conclusions and Future Work}

In this paper, inspired by the link functions in previous works, with
the support from Information Theory, we propose a new link function
of a text window, parameterized by the embeddings of words and the
residuals of bigrams. Based on the link function, we establish a generative
model of documents. The learning objective is to find a set of embeddings
maximizing their posterior likelihood given the corpus. This objective
is reduced to weighted low-rank positive-semidefinite approximation,
subject to Tikhonov regularization. Then we adopt a Block Coordinate
Descent algorithm, jointly with an online blockwise regression algorithm
to find an approximate solution. On seven benchmark sets, the learned
embeddings show competitive and stable performance.

In the future work, we will incorporate global latent factors into
this generative model, such as topics, sentiments, or writing styles,
and develop more elaborate models of documents. Through learning such
latent factors, important summary information of documents would be
acquired, which are useful in various applications.

\section*{Acknowledgments}

We thank Omer Levy, Thomas Mach, Peilin Zhao, Mingkui Tan, Zhiqiang
Xu and Chunlin Wu for their helpful discussions and insights. This
research is supported by the National Research Foundation, Prime Minister's
Office, Singapore under its IDM Futures Funding Initiative and administered
by the Interactive and Digital Media Programme Office.

\begin{appendices}\renewcommand\thesection{Appendix \Alph{section}}

\section{Possible Trap in SVD\label{sec:svdtrap}}

Suppose $\boldsymbol{M}$ is the bigram matrix of interest. SVD embeddings
are derived from the low rank approximation of $\boldsymbol{M}^{\T}\boldsymbol{M}$,
by keeping the largest singular values/vectors. When some of these
singular values correspond to negative eigenvalues, undesirable correlations
might be captured. The following is an example of approximating a
PMI matrix.

A vocabulary consists of 3 words $s_{1},s_{2},s_{3}$. Two corpora
derive two PMI matrices: 
\[
\boldsymbol{M}^{(1)}=\left(\begin{smallmatrix}1.4 & 0.8 & 0\\
0.8 & 2.6 & 0\\
0 & 0 & 2
\end{smallmatrix}\right),\quad\boldsymbol{M}^{(2)}=\left(\begin{smallmatrix}0.2 & -1.6 & 0\\
-1.6 & -2.2 & 0\\
0 & 0 & 2
\end{smallmatrix}\right).
\]

They have identical left singular matrix and singular values $(3,2,1)$,
but their eigenvalues are $(3,2,1)$ and $(-3,2,1)$, respectively. 

In a rank-2 approximation, the largest two singular values/vectors
are kept, and $\boldsymbol{M}^{(1)}$ and $\boldsymbol{M}^{(2)}$
yield identical SVD embeddings $\boldsymbol{V}=\left(\begin{smallmatrix}0.45 & 0.89 & 0\\
0 & 0 & 1
\end{smallmatrix}\right)$ (the rows may be scaled depending on the algorithm, without affecting
the validity of the following conclusion). The embeddings of $s_{1}$
and $s_{2}$ (columns 1 and 2 of $\boldsymbol{V}$) point at the same
direction, suggesting they are positively correlated. However as $\boldsymbol{M}_{1,2}^{(2)}=\boldsymbol{M}_{2,1}^{(2)}=-1.6<0$,
they are actually negatively correlated in the second corpus. This
inconsistency is because the principal eigenvalue of $\boldsymbol{M}^{(2)}$
is negative, and yet the corresponding singular value/vector is kept.

When using eigendecomposition, the largest two positive eigenvalues/eigenvectors
are kept. $\boldsymbol{M}^{(1)}$ yields the same embeddings $\boldsymbol{V}$.
$\boldsymbol{M}^{(2)}$ yields $\boldsymbol{V}^{(2)}=\left(\begin{smallmatrix}-0.89 & 0.45 & 0\\
0 & 0 & 1.41
\end{smallmatrix}\right)$ , which correctly preserves the negative correlation between $s_{1},s_{2}$.

\section{Information Theory}

\emph{Redundant information} refers to the reduced uncertainty by
knowing the value of any one of the conditioning variables (hence
redundant). \emph{Synergistic information} is the reduced uncertainty
ascribed to knowing all the values of conditioning variables, that
cannot be reduced by knowing the value of any variable alone (hence
synergistic).

\begin{figure}
\begin{centering}
\includegraphics[bb=0bp 35bp 283bp 210bp,scale=0.6]{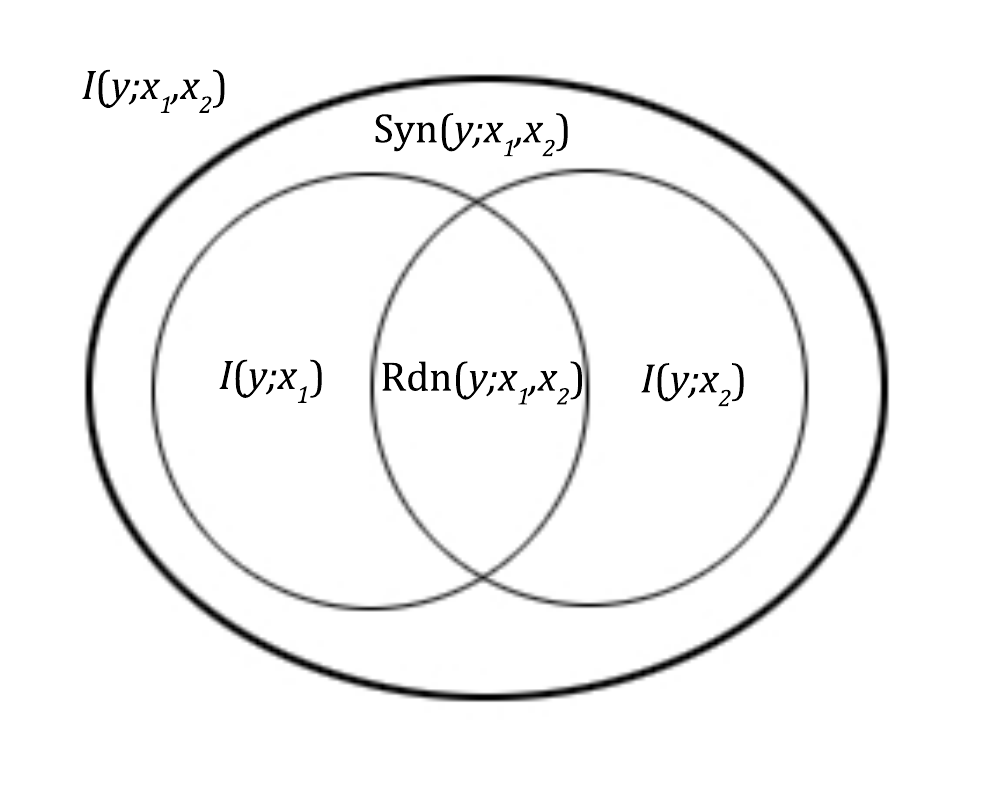}
\par\end{centering}

\centering{}\protect\caption{\label{fig:interaction-information}Different types of information
among 3 random variables $y,x_{1},x_{2}$. $I(y;x_{1},x_{2})$ is
the mutual information between $y$ and $(x_{1},x_{2})$. $\textrm{Rdn}(y;x_{1},x_{2})$
and $\textrm{Syn}(y;x_{1},x_{2})$ are the redundant information and
synergistic information between $x_{1},x_{2}$, conditioning $y$,
respectively.}
\end{figure}

The mutual information $I(y;x_{i})$ and the redundant information
$\textrm{Rdn}(y;x_{1},x_{2})$ are defined as: 
\begin{align*}
I(y;x_{i}) & =E_{P(x_{i},y)}[\log\frac{P(y|x_{i})}{P(y)}]\\
\negthickspace\textrm{Rdn}(y;x_{1},x_{2}) & =E_{P(y)}\negmedspace\left[\min_{x_{1},x_{2}}E_{P(x_{i}|y)}[\log\negmedspace\frac{P(y|x_{i})}{P(y)}]\right]
\end{align*}

The synergistic information $\textrm{Syn}(y;x_{1},x_{2})$ is defined
as the PI-function in \cite{multinfo}, skipped here.

The interaction information $\textrm{Int}(x_{1},x_{2},y)$ measures
the relative strength of $\textrm{Rdn}(y;x_{1},x_{2})$ and $\textrm{Syn}(y;x_{1},x_{2})$
\cite{infos}:
\begin{align*}
 & \textrm{Int}(x_{1},x_{2},y)\\
= & \textrm{Syn}(y;x_{1},x_{2})-\textrm{Rdn}(y;x_{1},x_{2})\\
= & I(y;x_{1},x_{2})-I(y;x_{1})-I(y;x_{2})\\
= & E_{P(x_{1},x_{2},y)}[\log\frac{P(x_{1})P(x_{2})P(y)P(x_{1},x_{2},y)}{P(x_{1},x_{2})P(x_{1},y)P(x_{2},y)}]
\end{align*}

Figure \ref{fig:interaction-information} shows the relationship of
different information among 3 random variables $y,x_{1},x_{2}$ (based
on Fig.1 in \cite{multinfo}).

PMI is the pointwise counterpart of mutual information $I$. Similarly,
all the above concepts have their pointwise counterparts, obtained
by dropping the expectation operator. Specifically, the \textit{pointwise
interaction information} is defined as $\textrm{PInt}(x_{1},x_{2},y)=\textrm{PMI}(y;x_{1},x_{2})-\textrm{PMI}(y;x_{1})-\textrm{PMI}(y;x_{2})=\log\frac{P(x_{1})P(x_{2})P(y)P(x_{1},x_{2},y)}{P(x_{1},x_{2})P(x_{1},y)P(x_{2},y)}$.
If we know $\textrm{PInt}(x_{1},x_{2},y)$, we can recover $\textrm{PMI}(y;x_{1},x_{2})$
from the mutual information over the variable subsets, and then recover
the joint distribution $P(x_{1},x_{2},y)$.

As the pointwise redundant information $\textrm{PRdn}(y;x_{1},x_{2})$
and the pointwise synergistic information $\textrm{PSyn}(y;x_{1},x_{2})$
are both higher-order interaction terms, their magnitudes are usually
much smaller than the PMI terms. We assume they are approximately
equal, and thus cancel each other when computing PInt. Given this,
PInt is always $0$. In the case of three words $w_{0},w_{1},w_{2}$,
$\textrm{PInt}(w_{0},w_{1},w_{2})=0$ leads to $\textrm{PMI}(w_{2};w_{0},w_{1})=\textrm{PMI}(w_{2};w_{0})+\textrm{PMI}(w_{2};w_{1})$.

\end{appendices}

\bibliographystyle{acl}
\nocite{*}
\bibliography{emnlp2015}

\end{document}